\documentclass[10pt, journal, twocolumn]{IEEEtran}

\usepackage{amsmath,amssymb}
\usepackage{graphicx}
\usepackage{bm}

\newcommand{\trp}{^\top}
\newcommand{\inv}{^{-1}}
\newcommand{\vx}{\mathbf{x}}
\newcommand{\vy}{\mathbf{y}}
\newcommand{\vw}{\mathbf{w}}
\newcommand{\vmu}{\boldsymbol{\mu}}
\newcommand{\vSigma}{\boldsymbol{\Sigma}}
\newcommand{\loss}{\mathcal{L}}
\newcommand{\braket}[2]{\ensuremath{{#1}\trp #2}}

\newcommand{\norm}[1]{\ensuremath{\Vert{#1}\Vert}}
\newcommand{\field}[1]{\ensuremath{\mathbb{#1}}}
\newcommand{\reals}{\field{R}}

\newcommand{\normal}{\mathcal{N}}
\newcommand{\MAP}{{\mbox{\tiny MAP}}}

\DeclareMathOperator*{\E}{\rm E}

\begin{document}
\title{Bayesian Extensions of Kernel Least Mean Squares}
\author{Il Memming Park, Sohan Seth, Steven Van Vaerenbergh\thanks{IP, SS, and SVV are postdoctoral researchers at University of Texas at Austin, USA, Helsinki Institute for Information Technology, Finland, and University of Cantabaria, Spain, respectively. The author list is in order of contribution.}}
\maketitle
\begin{abstract}
The kernel least mean squares (KLMS) algorithm is a computationally efficient nonlinear adaptive filtering method that ``kernelizes'' the celebrated (linear) least mean squares algorithm.
We demonstrate that the least mean squares algorithm is closely related to the Kalman filtering, and thus, the KLMS can be interpreted as an approximate Bayesian filtering method.
This allows us to systematically develop extensions of the KLMS by modifying the underlying state-space and observation models.
The resulting extensions introduce many desirable properties such as ``forgetting'', and the ability to learn from discrete data, while retaining the computational simplicity and time complexity of the original algorithm.
\end{abstract}

\section{Introduction}
Adaptive filtering algorithms deal with real-time learning scenarios, in which the environment is often nonstationary.
In general, these algorithms need to fulfill three basic requirements: 1) to sequentially learn from each observation; 2) to be adaptive to changing environments; and 3) to be computationally efficient.
Among many existing algorithms that fulfill these requirements, one that has stood the test of time is the celebrated least mean squares (LMS) algorithm. This algorithm has several interesting properties, in particular its inherent computational simplicity, and its implicit tracking ability despite its assumption of stationarity.

Inspired by the success of the LMS algorithm, a ``kernelization'' has been recently proposed under the name kernel least mean squares (KLMS) algorithm~\cite{Liu2008a}.
The KLMS inherits many desirable properties of LMS and extends it to a large class of nonlinear filtering algorithms.
Nevertheless, it has certain limitations that arise from its formulation as an adaptive filter in a possibly infinite dimensional feature space.
Specifically, if implemented naively, the representation of the filter grows linearly with the number of data samples processed.
Moreover, both the LMS and the KLMS explicitly minimize the squared error between the desired and the estimated observation values, hence, they cannot be naturally applied to problems with discrete observations such as class labels.
To summarize, KLMS, in its current format,
\begin{enumerate}
\item cannot be extended to tackle discrete observations,
\item grows indefinitely with new observations, and
\item does not provide an explicit understanding of its tracking ability (see below).
\end{enumerate}

\begin{figure}[t!]
\centering
\includegraphics[width=3.2in]{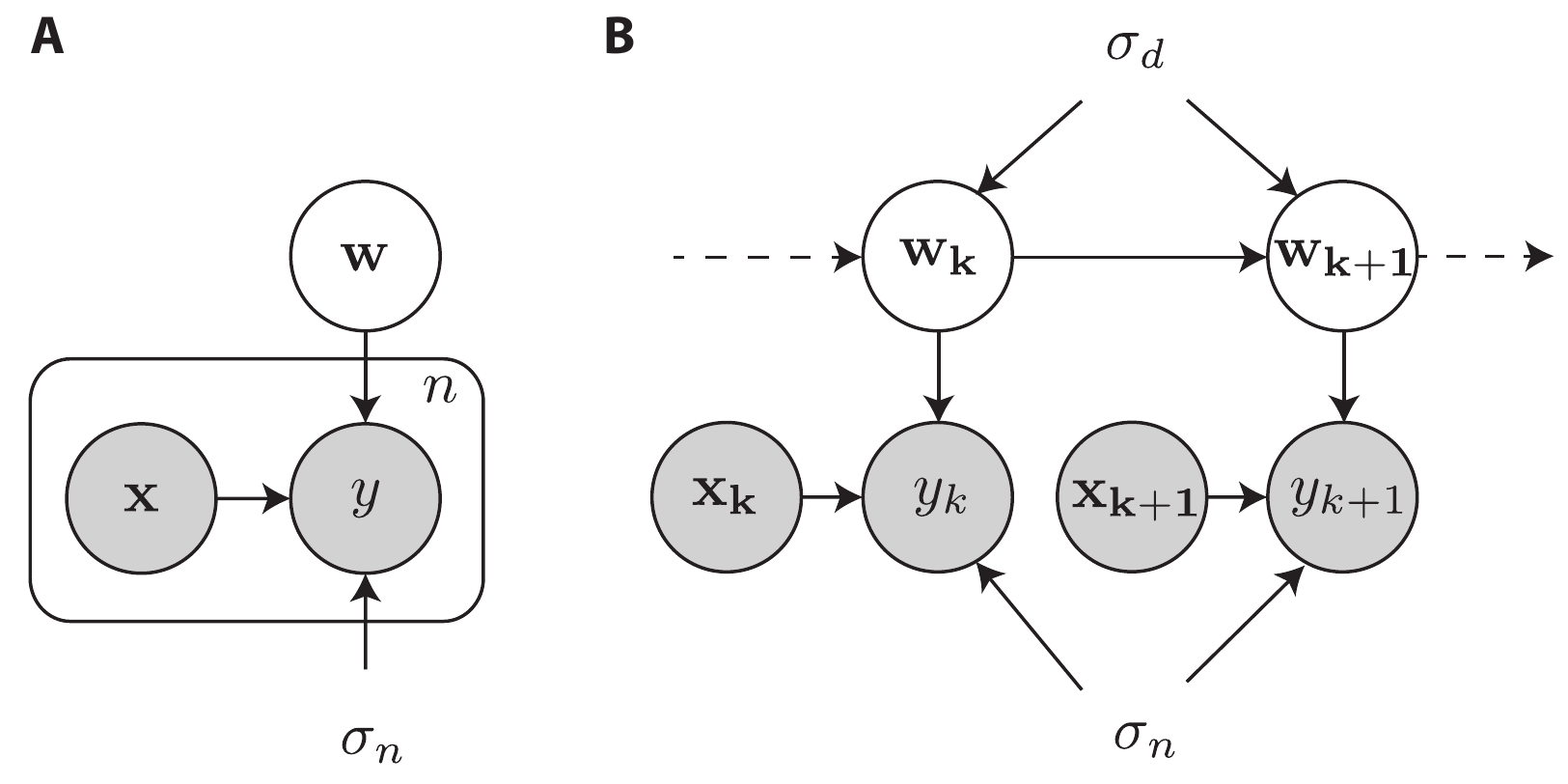}
\caption{Graphical models illustrating the contrast between stationary and nonstationary generative processes.
The arrows signify the conditional dependence between variables.
Gray shaded circles denote observed variables, and the box denotes repetition.
(A) Stationary model.
Each observation pair $(x_i,y_i)$ is assumed to have the same relation $\vw$ as in the classic regression setting.
Original derivation of LMS is given in this context.
(B) Non-stationary model.
The weight $\vw$ evolves over time---hence the relation between $(x_i,y_i)$---through diffusion with parameter $\sigma_d$.
The Kalman filter is derived under this model.
We show the KLMS can also be derived from the nonstationary model.
}
\label{fig:graphical:model}
\end{figure}

The available adaptive filters in the statistical signal processing literature can be broadly categorized as 1) filters derived from a (stationary) regression framework, such as LMS and recursive least squares (RLS), and 2) filters derived from a (nonstationary) state-space framework such as the Kalman filter (Fig.~\ref{fig:graphical:model}).
These two classes of algorithms are both successfully applied to nonstationary systems, and occasionally show similar computational characteristics, e.g., extended RLS can be estabilished as a special case of the Kalman filter~\cite{Sayed2003}.
We seek a principled explanation---based on the Bayesian filtering framework---of how LMS achieves tracking despite its formulation under the assumption of stationarity, since it would allow us to systematically address the limitations of the KLMS discussed earlier.
The connection we seek is related but distinct from that of Bayesian approaches recently explored by~\cite{vanvaerenbergh2012kernel} for kernel RLS (KRLS);
specifically, it was shown that a recursive filtering implementation of the Bayesian regression framework (Gaussian processes) naturally leads to a formulation that is equivalent to kernelization of RLS.
They have explicitly introduced two types of forgetting methods to enable tracking---from which we were inspired for extending the KLMS in this paper---but the forgetting is not incorportated within the Bayesian framework.

In this paper, we show that LMS (and KLMS) can indeed be derived as an approximation of a state-space based Bayesian filtering (section \ref{sec:bayes}).
In order to achieve a low computational complexity, though, only the mode of the posterior distribution can be estimated and retained for each sample.
This new interpretation allows us to derive extensions of the KLMS algorithm by tweaking the underlying state-space and observation models (section \ref{sec:forgetting} and \ref{sec:observation}).
Here, we extend the KLMS algorithm to integer and binary-valued observations, and also introduce a forgetting factor in order to improve its tracking ability.
We furthermore illustrate each extension with a simple example to demonstrate its performance in tracking nonstationary signals, and compare with existing methods such as the quantized KLMS (QKLMS)~\cite{Chen2012Quantized} and naive online regularized risk minimization algorithm (NORMA)~\cite{Kivinen2004}.
Some interesting properties of these extensions are that each algorithm learns by assigning a coefficient for each new observation, and that the time complexity of the $n$-th iteration has linear cost, $\mathcal{O}(n)$, in both space and time, as is the case for the KLMS.

\renewcommand{\H}{\mathcal{H}}
\section{Stochastic gradient descent derivation} \label{sec:lms}
Both Widrow \& Hoff's LMS, and KLMS are derived from mean squared error cost function~\cite{Haykin2002,Kivinen2004,Liu2008a,Liu2010} which is prevalent in traditional signal processing.
The filtering setting assumes a linear model
$$
f(\vx;\vw) = \braket{\vw}{\phi(\vx)}
$$
in the \emph{feature space} where $\phi(\vx) \in \H$ is the feature vector associated with the input vector $\vx$, and $\vw \in \H$ is the vector representation of the filter in a Hilbert space $\H$.
The following derivation holds for both LMS and KLMS, taking into account that for LMS the feature space is the (Euclidean) input space itself, i.e., $\phi(\vx) = \vx \in \reals^d$, while KLMS uses a (potentially) infinite dimensional (reproducing kernel) Hilbert space induced by a positive definite kernel $k: \reals^d \times \reals^d \to \reals$ where $k(\vx,\vy) = \braket{\phi(\vx)}{\phi(\vy)}$~\cite{Scholkopf2002}.

The mean squared error is defined as,
\begin{align}
    \loss_{\mbox{mse}}(\vw) =
	\frac{1}{2}
	\E[
	(f(\mathbf{X};\vw) - \mathbf{Y})^2
	],
\end{align}
where $\mathbf{X}$ and $\mathbf{Y}$ are the random vector and variable for the input signal and the desired output, respectively.
Note that this cost function corresponds to a regression problem corrupted by Gaussian noise assuming independence, illustrated in  Fig.~\ref{fig:graphical:model}A, where
given a set of $n$ input-output data pairs $\mathcal{D}_n = ( ( \vx_1, y_1), \ldots, (\vx_n, y_n) )$, the negative log-likelihood is given by
\begin{align}
    \loss_A(\vw) = \frac{1}{n}\sum_{i=1}^n \frac{1}{2 \sigma_n^2}
    \left(
	f(\vx_i;\vw) - y_i
    \right)^2 + \mathrm{const}
\end{align}
where $\sigma_n^2$ is the noise variance.
In the limit of large number of \textit{iid} samples, due to the law of large numbers, it converges to its expectation which is the mean squared error (up to a factor and a constant).
Therefore, the (asymptotic) maximum likelihood solution under this model coincides with the minimum MSE (MMSE) solution.

The basic steepest descent learning rule has the form
\begin{align}\label{eq:steepest}
    \Delta \vw \leftarrow -\eta \frac{\partial \loss_{\mbox{mse}}(\vw)}{\partial \vw}
    = -\eta \E\left[
	    (\braket{\vw}{\phi(\mathbf{X})} - \mathbf{Y})
	    \phi(\mathbf{X})
	\right].
\end{align}
Since the cost function is convex, it will converge to the MMSE solution for a sufficiently small learning rate $\eta$.
To make an online learning rule, a stochastic gradient descent is used in practice.
In particular, the learning rule of the LMS algorithm is obtained by dropping the expectation from~\eqref{eq:steepest}, which yields
\begin{align}\label{eq:KLMS:adhoc}
    \vw_{i+1} \leftarrow \vw_i
	-\eta_i (\braket{\vw_i}{\phi(\vx_i)} - y_i)
	\phi(\vx_i).
\end{align}
Hence, after processing $i$ samples, the prediction for the next sample $y_{i+1}$ is given by,
\begin{align}
    \hat{y}_{i+1}
    	= \braket{\vw_{i}}{\phi(\vx_{i+1})}
    	= \sum_{k=1}^i \eta e_k \braket{\phi(\vx_k)}{\phi(\vx_{i+1})}
\end{align}
where $e_k = {y_k - \braket{\vw_{k}}\phi(\vx_k)}$ is the error for each sample.
For KLMS, the prediction can be directly computed from the samples despite $f \in \H$, since $\braket{\phi(\vx_i)}{\phi(\vx_k)} = k(\vx_i, \vx_k)$.

The stochastic gradient descent algorithm is guaranteed to convergence (almost surely) to the global optimal solution under stationary and ergodic observations
if a proper step size scheduling is used (e.g., $\sum_{i=1}^\infty \eta_i^2 < \infty$ and $\sum_{i=1}^\infty \eta_i = \infty$).
However, the tracking capability of LMS/KLMS is dependent on the step size; if the step-size were annealed, it would be tracking less efficiently as more samples are seen.
Therefore, to have constant tracking, step size is not annealed in practice, that is, $\forall i \; \eta_i = \eta$.
Then, for the price of non-zero misadjustment, the algorithms can surprisingly learn continuously from new samples, and overwrite what was learned before.
Note that this ``hack'' disconnects the algorithm from the graphical model Fig.~\ref{fig:graphical:model}A which inherently assumes a stationary data generation process.
In the following section, we show how this tracking ability can be derived from first principles.

\section{Bayesian interpretation} \label{sec:bayes}
\subsection{Model}
A slowly changing system can be explicitly described by a probabilistic model with latent dynamics on the parameter.
In such a model, each parameter associated with each sample or time is considered as an interdependent random (latent) variable, as illustrated in Fig.~\ref{fig:graphical:model}B.
Our goal is to show that the KLMS is an approximate sequential inference for $\vw_i$.
We start with a diffusion process as a reasonable model for nonstationarity:
\begin{align}\label{eq:diffusion:pure}
    P(\vw_{k+1}|\vw_k) &= \mathcal{N}(\vw_{k+1}; \vw_k, \sigma_d^2 \mathbf{I}),
\end{align}
where $\mathcal{N}$ denotes a Gaussian distribution in the feature space, and $\sigma_d^2$ is the variance of diffusion on each direction.
The likelihood model is assumed to be a linear--Gaussian model, similar to the stationary case,
\begin{align}\label{eq:likelihood:gaussian}
    P(y_k|\vx_k, \vw_k) &= \mathcal{N}(y_k; \vw_k\trp \phi(\vx_k), \sigma_n^2)
\end{align}
where $\sigma_n^2$ is the observation noise variance.
We remark that the conditional distributions \eqref{eq:diffusion:pure} and \eqref{eq:likelihood:gaussian} for the finite dimensional feature space is a special case of the Kalman filter model with linear dynamics.

\subsection{Approximate inference}
If we wish to (recursively) infer the posterior weight distribution $p(\vw_k | \mathcal{D}_k)$, we only need the mean and covariance, since we assume Gaussianity in this model~\cite{Roweis1999}.
Assuming $P(\vw_{k-1}|\mathcal{D}_{k-1}) = \mathcal{N}(\vmu_{k-1}, \vSigma_{k-1})$, a single linear Gaussian observation results in a one step evolution of the posterior as another Gaussian $P(\vw_k|\mathcal{D}_k) = \mathcal{N}(\vmu_k, \vSigma_k)$, with
\begin{align}
    \vSigma_k\inv &=
	\vSigma_{k-1}\inv + \frac{1}{\sigma_n^2} \phi(\vx_k)\phi(\vx_k)\trp
    \\
    \vmu_k &= \vSigma_k
	\left[
	    \frac{1}{\sigma_n^2} y_k \phi(\vx_k) + \vSigma_{k-1}\inv \vmu_{k-1}
	\right].
\end{align}
This recursion can be solved efficiently, and the solution is known as the
extended recursive least squares algorithm~\cite{liu2009extended}. However,
it requires a quadratic number of operations in terms of the dimension of the feature vector for updating the (inverse) covariance matrix.
In case of an infinite-dimensional feature space, the feature vector dimension grows linearly with the number of observations, rendering this approach prohibitive.
Therefore, in order to obtain a linear time complexity algorithm, we assume the posterior to be concentrated around the maximum.
In other words, we approximate the posterior
as a delta function at the maximum a posteriori (MAP) estimate
$P(\vw_{k}|\mathcal{D}_k) \simeq \delta_{\vw_k^{\mbox{\tiny MAP}}}$
before inferring $P(\vw_{k+1}|\mathcal{D}_k)$.
Below, we show the steps for online inference rules using this approximation.

First, the approximation is equivalent to assuming an isotropic Gaussian around the MAP estimate for the previous sample.
\begin{align}
    P(\vw_{k+1}|\mathcal{D}_k)
	&= \int P(\vw_{k+1}|\vw_k) P(\vw_k|\mathcal{D}_k) \mathrm{d}\vw_k
	\nonumber
	\\
	&\simeq \int P(\vw_{k+1}|\vw_k) \delta(\vw_k^{\mbox{\tiny MAP}}) \mathrm{d}\vw_k
	\nonumber
	\\
	&= P(\vw_{k+1}|\vw_k^{\mbox{\tiny MAP}})
	= \mathcal{N}(\vw_k^{\mbox{\tiny MAP}}, \sigma_d^2 \mathbf{I}).
	\label{eq:approximate:diffusion}
\end{align}
Using Bayes' rule, the posterior weight distribution is,
\begin{align}
    P(\vw_{k+1}|\mathcal{D}_{k+1}) &\propto
	P( y_{k+1} | \vx_{k+1}, \vw_{k+1}) P(\vw_{k+1} | \mathcal{D}_k)
	\nonumber
	\\
	&=
	\mathcal{N}(y_{k+1}; \vw_{k+1}\trp \phi(\vx_{k+1}), \sigma_n^2)
	\nonumber
	\\
	& \qquad \cdot
	\mathcal{N}(\vw_{k+1}; \vw_k^{\mbox{\tiny MAP}}, \sigma_d^2 \mathbf{I})
	\nonumber
	\\
	&=
	\mathcal{N}(\vw_{k+1}; \vw_{k+1}^\MAP, \vSigma_{k+1})
\end{align}
where the parameters for the posterior are,
\begin{align}
    \vSigma_{k+1}\inv &= \frac{1}{\sigma_d^2}\mathbf{I}
	+ \frac{1}{\sigma_n^2} \phi(\vx_{k+1})\phi(\vx_{k+1})\trp
    \\
    \vw_{k+1}^\MAP
    &= \vSigma_{k+1}
	\left[
	    \frac{\vw_k^{\mbox{\tiny MAP}}}{\sigma_d^2}
	    + \frac{y_{k+1}\phi(\vx_{k+1})}{\sigma_n^2}
	\right].
\end{align}
This can be simplified using the matrix inversion lemma,
\begin{align}
\vw_{k+1}^\MAP
    &= \vw_k^\MAP +
	\frac{\eta' (y_{k+1}
	    - \braket{\vw^{\MAP}_k}{\phi(\vx_{k+1}}))\phi(\vx_{k+1})
	}{
	    1 + \eta' \braket{\phi(\vx_{k+1})}{\phi(\vx_{k+1})}
	}
\label{eq:wmap1}
\end{align}
where the learning rate is determined by the diffusion-to-noise ratio
$\eta' = \sigma_d^2 / \sigma_n^2$.
This is very similar to the normalized LMS (NLMS) update rule, although not identical.
If the kernel is normalized, such that $k(\vx, \vx) = 1$, then it can be simply rewritten as,
\begin{align}\label{eq:KLMS}
    \vw_{k+1}^\MAP &= \vw_k^\MAP + \eta e_k \phi(\vx_k)
\end{align}
where $\eta = \eta' / (1+\eta')$.
Note that the stochastic gradient derivation \eqref{eq:KLMS:adhoc} is identical to the approximate Bayesian learning rule \eqref{eq:KLMS}; we have rederived KLMS with a state-space model.
Also, note that $0 < \eta < 1$, thus we have a frequentist convergence guarantee of the weight vector to the optimal weight vector $\vw^\ast$ in mean in a stationary environment, i.e., $\lim_{k\to\infty}\E[\vw_k] \rightarrow \vw^\ast$~\cite{Liu2008a}.

\subsection{Implications}
The resulting algorithm~\eqref{eq:KLMS} has linear time complexity $\mathcal{O}(m)$ where $m$ is the number of $\vx_i$'s used to represent the weight, while the extended KRLS requires quadratic time complexity $\mathcal{O}(m^2)$ at each iteration~\cite{liu2009extended,vanvaerenbergh2012kernel}.
However, this does not come without a sacrifice.
This derivation forgets the covariance of the weights after each iteration.
The covariance contains the information on the uncertainty over the weight vector; the weights with less uncertainty are updated less.
However, with the covariance being approximated by a constant diagonal~\eqref{eq:approximate:diffusion}, roughly speaking, the weight vector is updated equally, and independently.
This brings a couple of disadvantages.
First, the posterior broadness information is lost, so we cannot report how much confidence we have about the current estimate.
Second, the update is not optimal, and we cannot guarantee its asymptotic convergence to the Bayesian posterior.
However, as we will see in the following section, we have some practical frequentist convergence results.
\newcommand{\vq}{\mathbf{q}}
\subsection{Tracking}
Despite the possible slower convergence, we provide a frequentist guarantee that KLMS tracks the true weight under mild conditions.
Specifically, we assume that the true system $f^\ast(\vx) = {\vw^\ast}\trp\phi(\vx)$ slowly changes over time, i.e.,
$\vw^\ast_{k+1} = \vw^\ast_k + \vq_k$ where $\vq_k \in \H$ is a small independent stochastic perturbation with $\E[\mathbf{q}] = 0$ and $\E[ \mathbf{q}\trp \mathbf{q} ] = \sigma^2_q$.
From~\eqref{eq:KLMS}, we can write the difference in the estimate as,
\begin{align}
    \Delta \vw_{k+1} &:= \vw_{k+1}^\ast - \vw_{k+1}
	=\vw_k^\ast + \vq_k -\vw_k - \eta e_k \phi(\vx_k)
    \nonumber \\
	&= \Delta\vw_k + \vq_k - \eta e_k \phi(\vx_k)
    .\label{eq:prime}
\end{align}
Our goal is to bound the asymptotic expected norm of~\eqref{eq:prime}.
The norm can be expanded as,
\begin{align}
    \norm{\Delta\vw_{k+1}}^2
	&=
	\norm{\Delta\vw_k}^2 + \norm{\vq_k}^2 + \eta^2 e_k^2 \braket{\phi(\vx_k)}{\phi(\vx_k)}
\nonumber \\
&\quad
+ 2\braket{\vq_k}{\Delta\vw_k} - 2\eta e_k \braket{\vq_k}{\phi(\vx_k)}
\nonumber \\
&\quad
- 2\eta e_k \braket{\Delta\vw_k}{\phi(\vx_k)}.
\end{align}
By taking the expectation on both sides, we get,
\begin{align*}
    \E \norm{\Delta\vw_{k+1}}^2
    &= \E \norm{\Delta\vw_k}^2 + \sigma^2_q + \eta^2\E [e_k^2]
    - 2\eta \E[e_k \zeta_k]
\end{align*}
where $\zeta_k = \left(\vw_k^\ast - \vw_k\right)\trp\phi(\vx_k)$ is the model mismatch error, and we have assumed $k(\vx,\vx) = 1$ for simplicity. Notice that the error $e_k$ can be decomposed as $e_k = \nu_k + \zeta_k$ where $\nu_k$ and $\zeta_k$ are independent of each other, and also $\nu_k \sim \mathcal{N}(0, \sigma_n^2)$. Therefore, by taking the limit $k \rightarrow \infty$, we obtain the steady state condition:
\begin{align} \label{eq:steady}
    &\sigma^2_q + \eta^2 \E [\zeta_\infty^2]
    + \eta^2 \sigma_n^2
    - 2\eta\E [\zeta_\infty^2] = 0.
\nonumber\\
\Rightarrow &\E [\zeta_\infty^2]
    = \frac{\sigma^2_q  + \eta^2 \sigma_n^2}{\eta(2-\eta)}.
\end{align}
In a stationary environment we have $\sigma^2_q = 0$, and therefore, the steady state error disappears as $\eta$ tends to zero.
However, in a nonstationary environment such is not the case, since in the limiting learning rate value the filter fails to track.
Thus, for tracking, one needs a finite learning rate value.
A suitable learning rate interval that guarantees convergence remains to be explored.

\section{Forgetful dynamics for KLMS}\label{sec:forgetting}
We made a theoretical connection between the KLMS and the Kalman filter in section \ref{sec:bayes}, providing a state-space interpretation to KLMS which explains why KLMS is appropriate for nonstationary environments.
In this section, we extended them in a principled manner to obtain forgetting dynamics in the weights.
This is achieved by a simple modification of the latent dynamics \eqref{eq:diffusion:pure}; we will generalize the noise distribution (likelihood function) \eqref{eq:likelihood:gaussian} in section~\ref{sec:observation}.

Instead of a pure random walk dynamics~\eqref{eq:diffusion:pure}, we add a leakage towards the origin, effectively forgetting the past exponentially.
The resulting diffusion is a discrete-time analogue of the Ornstein-Uhlenbeck process (equivalently, a first order auto-regressive process).
\begin{align}\label{eq:OU}
    p(\vw_{k+1}|\vw_k) = \normal(\lambda\vw_k,\sigma_d^2).
\end{align}
The asymptotic marginal distribution of the prior dynamics is $\mathcal{N}(0, \sigma_d^2 / (1-\lambda^2) \mathbf{I})$, hence the weights are isotropically distributed around the origin in the absence of observation.
If $k(\vx,\vx)$ is constant, the RKHS norm is proportional to the function norm, and the norm of the corresponding functions follows a Gaussian distribution centered around the origin.
As a result the learning rule \eqref{eq:wmap1} becomes
\begin{align}\label{eq:fKLMS}
    \vw_{k}^\MAP = \lambda\vw_{k-1}^\MAP + \frac{\eta (y_{k} - \lambda\vw^{\MAP\top}_{k-1}\phi(\vx_{k}))}{1+\eta \norm{\phi(\vx_{k})}^2} \phi(\vx_{k}).
\end{align}
This learning rule~\eqref{eq:fKLMS} is very similar to that of NORMA\footnote{A subtle yet notable difference between these two methods is that for NORMA, the error at the current location is computed before the  weights have been decayed.}, which aims to regularize the solution~\cite{Kivinen2004}.

\begin{figure}[t!]
\centering
\includegraphics[width=\columnwidth]{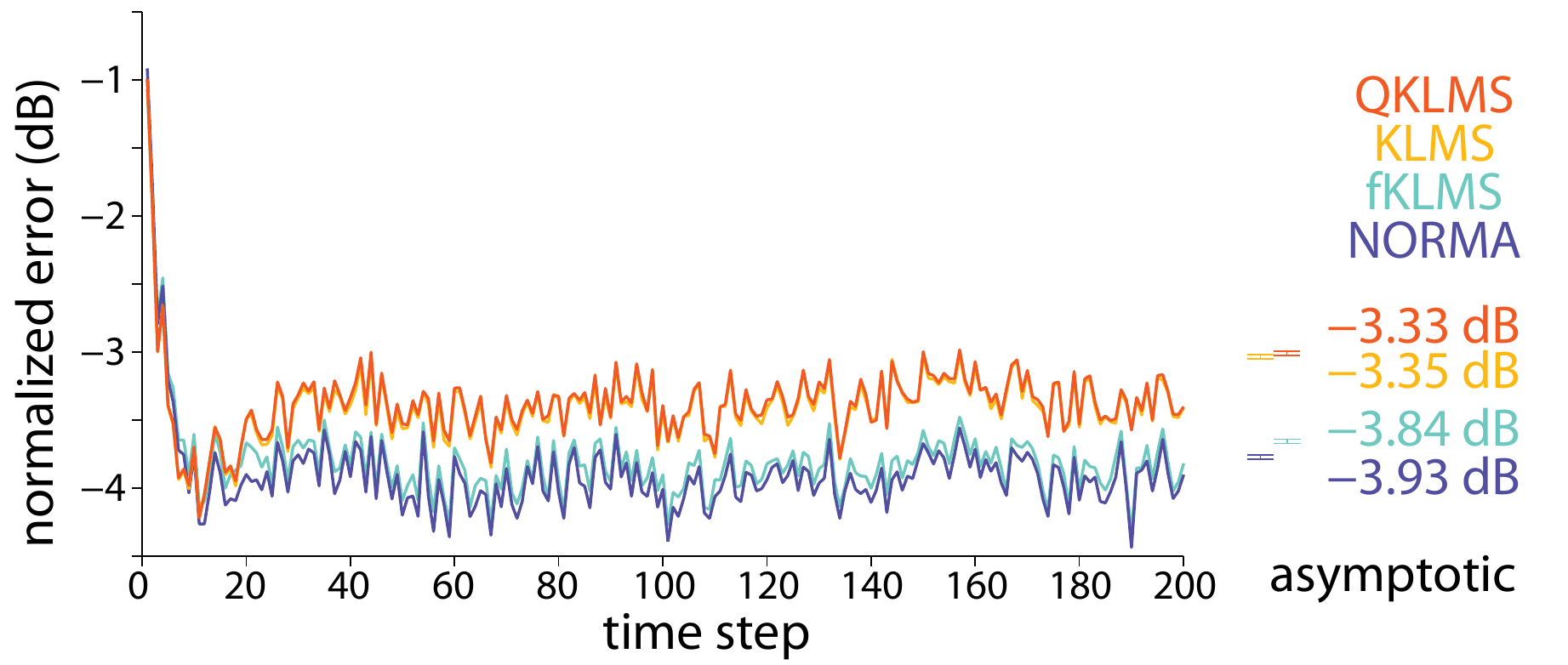}
\caption{
Comparison of KLMS-like algorithms for tracking non-stationary data.
The proposed forgetting dynamics extension of KLMS is denoted as fKLMS.
The data is generated as a realization of a spatio-temporal Gaussian process with covariance
$\exp({-(t-s)^2/2/10^2}) \exp({-(x-y)^2/2/0.2^2})$, where $s$ and $t$ indicate temporal indices, with $10$ dB independent additive Gaussian noise.
The budgets of the algorithms are fixed as follows: NORMA and fKLMS use a budget of $20$ centers, QKLMS has a budget parameter of $0.05$, yielding a maximum of $20$ centers in $[0,1]$, and KLMS is not constrained at all.
Pruning is accomplished in fKLMS and NORMA by dropping the oldest coefficient.
A squared exponential kernel with kernel size of $0.2$ is used for all the algorithms.
The MSE is computed for a one-step prediction, and averaged out over $2000$ simulations.
We scan several values of the parameters, and report the parameter with least average normalized MSE over the iterations.
Asymptotic normalized MSE is estimated from 800 samples after time step 200 (error bars indicate 2 standard error).
Interestingly, we observe that KLMS and QKLMS, which lack a forgetting mechanism, obtain weaker tracking performance.
}
\label{fig:tracking:fKLMS}
\end{figure}

Note that the learning rule~\eqref{eq:fKLMS} can be expanded, and rewritten as,
\begin{align}
    \vw_{k}^\MAP = \sum_{i=1}^k \lambda^{k-i} \beta_i \phi(\vx_{i}),
\end{align}
where $\beta_i$ is a scalar corresponding to the coefficient at the learning step.
We can see that each effective coefficient $\lambda^{k-i} \beta_i$ for each $\phi(\vx_i)$ shrinks geometrically over time.
Thus, the effect of older observation to the current weight estimate is small in general.
Note, however, that the algorithm forgets not by making the covariance larger as in the Kalman filter, but by changing the the mean.

Like NORMA, the forgetting dynamics KLMS can be interpreted as introducing a regularization for the weight vector.
However, as shown in \cite{Liu2010}, KLMS is self-regularizing and hence such extra regularization is usually not necessary.
However, it provides a significant practical benefit for maintaining a compact representation---we can prune the $\phi(\vx_i)$ component that are too small to have any effect.
A simple strategy of pruning is to drop effective coefficients below a small pre-specified threshold.
Then, assuming a stationary distribution over the new coefficients, the forgetting factor $\lambda$ and the threshold control the expected representation length.
One can also maintain a fixed budget by removing the oldest coefficient after a predefined number of centers is reached.
We employ the latter strategy in the experiment shown in Fig.~\ref{fig:tracking:fKLMS}.

\section{Novel observations models for KLMS}\label{sec:observation}
Gaussian observation model~\eqref{eq:likelihood:gaussian} is widely used for continuous observations, however, it is inappropriate where the observations are natural numbers, or binary labels.
In this section, we extend KLMS by replacing the Gaussian observation model with other distributions in the exponential family.

\subsection{Poisson observations}
Poisson likelihood is widely used when the observations are natural numbers: $0, 1, 2, \cdots$.
For example, in neuroscience, neural response is often quantified by the number of spikes, and tracking how the neural code changes during experiment is of great importance~\cite{Brown2001}.
We use the canonical inverse link function (exponential) for the Poisson distribution to map the linear (or nonlinear) function from the input to the non-negative rate parameter, i.e.,
\begin{align}\label{eq:likelihood:poisson}
    P(y_k|\vx_k, \vw_k) &= \mathrm{Poisson}(y_k; \exp(\vw_k\trp \phi(\vx_k))).
\end{align}
To derive the adaptive filtering algorithm, once again, we approximate the current state given the previous observations as
\eqref{eq:approximate:diffusion}, for which the log prior is,
\[
\log P(\vw_k | \mathcal{D}_{k-1}) =
-\frac{1}{2\sigma^2_d} (\lambda\vw^{\MAP}_{k-1} - \vw_k)\trp (\lambda\vw^{\MAP}_{k-1} - \vw_k)
+ c
.
\]
Therefore, using Bayes' rule, the posterior at time $k$ is,
\begin{align}
    \log P(\vw_k | \mathcal{D}_k) &= \log P(y_k|\vx_k, \vw_k) + \log P(\vw_k | \mathcal{D}_{k-1})
    \nonumber\\
     &= y_k \vw_k\trp \phi(\vx_k) - \exp(\vw_k\trp \phi(\vx_k))
    \nonumber\\
     &\quad - \frac{1}{2\sigma^2_d} (\lambda\vw^{\MAP}_{k-1} - \vw_k)\trp (\lambda\vw^{\MAP}_{k-1} - \vw_k),
    \nonumber
\end{align}
where irrelevant constants are omitted. We need to maximize this log-posterior over $\vw_k$ to estimate $\vw^{\MAP}_k$.
The MAP estimate must satisfy, the stationary condition ${\partial \log P(\vw_k|\mathcal{D}_k)}/{\partial \vw_k} = 0$, which implies,
\begin{align}
    2\sigma^2_d(y_k - \exp({\vw_k^\MAP}\trp \phi(\vx_k))) \phi(\vx_k) = (\lambda\vw^{\MAP}_{k-1} - \vw_k^\MAP).
    \label{eq:Poisson:stationary}
\end{align}
We observe that the solution of \eqref{eq:Poisson:stationary} can be expressed as,
\[
    \vw_k^\MAP = \lambda\vw^{\MAP}_{k-1} + \alpha_k\phi(\vx_k),
\]
where $\alpha_k$ is a scalar, and therefore, we can rewrite the log-posterior as,
\begin{align}
J(\alpha_k) = y_k(\log(\psi_{k}) + \alpha_k) - \psi_{k}\exp(\alpha_k) - \frac{\alpha_k^2}{2\sigma^2_d}
\end{align}
where $\psi_{k}=\exp(\lambda{\vw^{\MAP}_{k-1}}\trp \phi(\vx_k))$,
and we have assumed a normalized kernel for simplicity.
Thus, we reduce the problem of finding an infinite dimensional weight vector to a one dimensional optimization.
Although, there is no analytical solution, the cost function $J(\alpha_k)$ is strictly concave and therefore, its maximum can be easily found by existing optimization tools.
The complexity of this algorithm is still $\mathcal{O}(n)$, with a constant overhead for solving a concave maximization problem at each step.

\begin{figure}[t]
\includegraphics[width=\columnwidth]{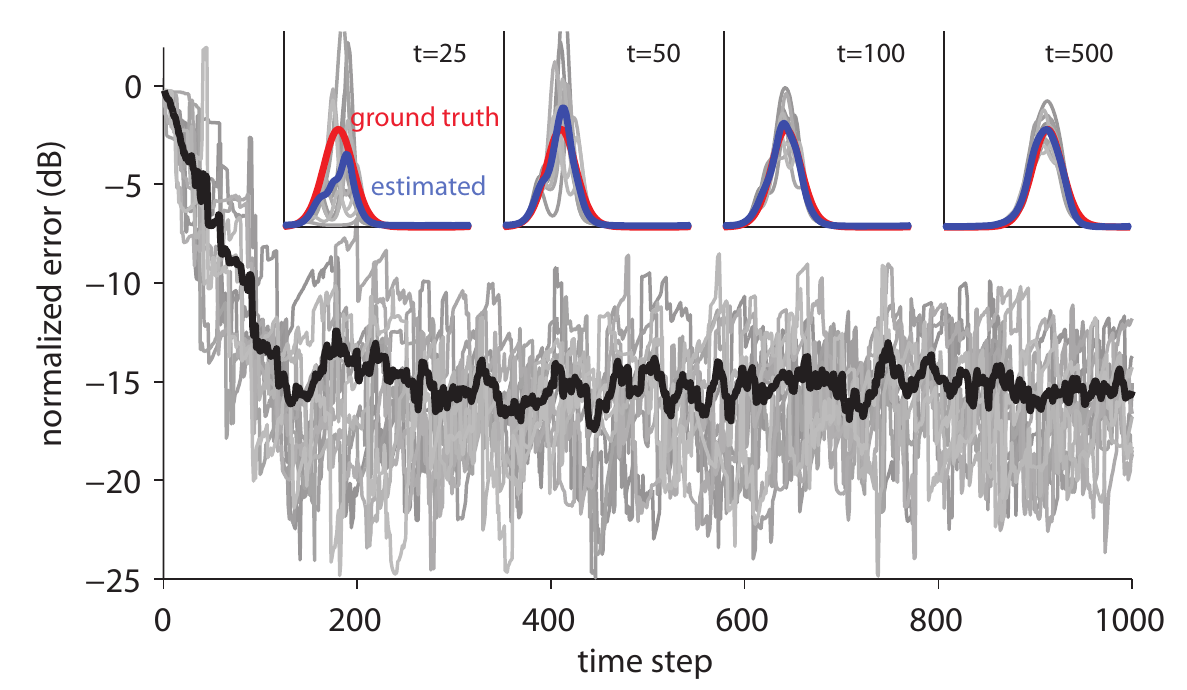}
\caption{Tracking example of the Poisson extension of KLMS algorithm.
 The observations model a slowly drifting tuning curve of a simple cell in V1. The tuning curve is modeled as an exponentiated cosine $\lambda(x) =\exp({4 \cos(x - \mu) - 0.1})$ where $\mu$ constantly drifted $100$ degrees during the $1000$ iterations.
    We measure the normalized estimation error between the true tuning curve and the estimated curve.
    Insets show the actual function estimate at 25, 50, 100, 500 time steps.
    Gray lines show 11 repeats of the experiment, and the dark curves correspond to their average.
    The kernel was $k(x,y) = \exp(-(x-y)^2/100)$ and $\sigma^2_d =0.1$.
}
\label{fig:tracking:poisson}
\end{figure}

We demonstrate its performance on a neurally inspired example in Fig.~\ref{fig:tracking:poisson}.
A typical nonlinear response function (tuning curve) is set to shift its center slowly, creating a non-stationary tracking problem.
The Poisson-KLMS extension correctly tracks, and maintains a small MSE throughout the experiment.

\subsection{Binary observations}
\begin{figure}[t!]
\includegraphics[width=\columnwidth]{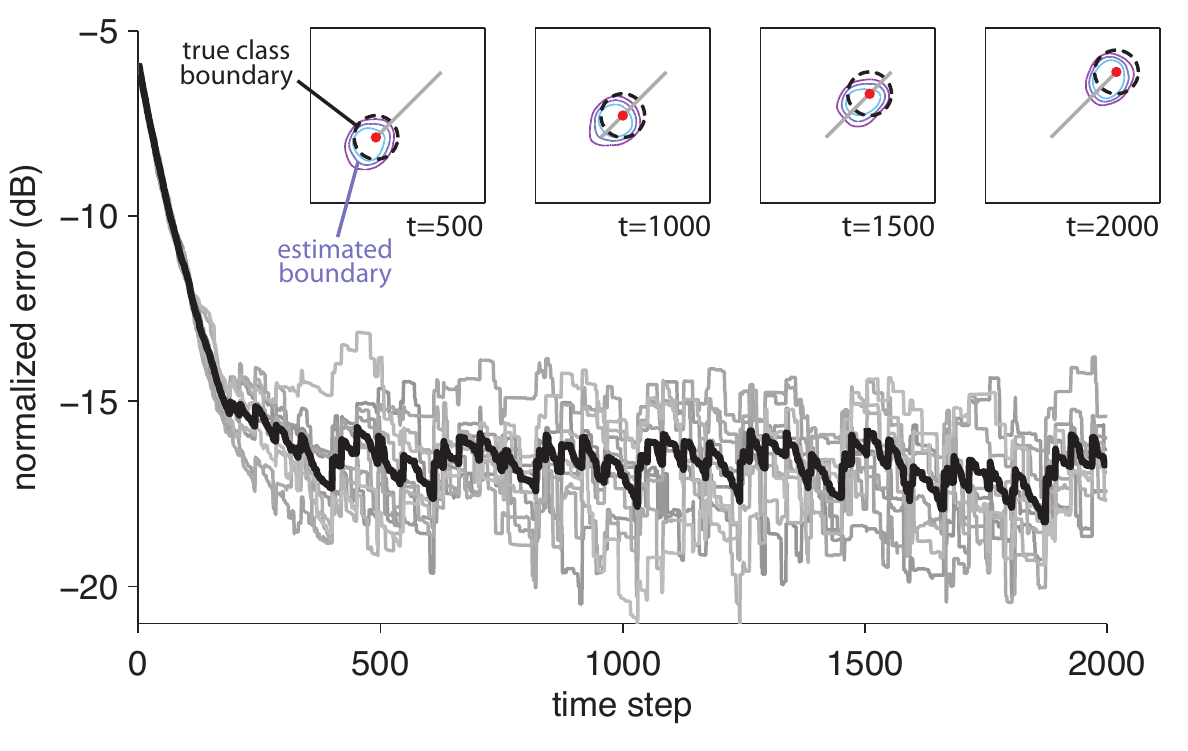}
\caption{Tracking example of the Logistic extension of KLMS algorithm.
    Insets:
    A two dimensional circular (radius 0.5) binary classification boundary is translated
    over time (gray line shows the trajectory of the center from $(-1,-1)$ to $(1,1)$).
    Three contours represent 0.25, 0.5, and 0.75 probability of the posterior.
    The kernel was $k(x,y) = \exp(-(x-y)^2/0.1)$ and $\sigma^2_d =6$.
}
\label{fig:tracking:logistic}
\end{figure}

Similarly, Bernoulli likelihood is widely used when the observations are binary labels, such as in a classification problem. Here we only address the binary classification problem, while the generalization to multi-class classification is certainly feasible, and straightforward.
We use the inverse canonical link function (logistic) for the Bernoulli distribution to map the linear (or nonlinear) function from the input to the non-negative probability score between $[0,1]$, i.e.,
\begin{align}\label{eq:likelihood:bernoulli}
P(y_k|\vx_k,\vw_k) = \mathrm{Bernoulli}(y_k;\mathrm{logistic}(\vw_k^\top\phi(\vx_k)))
\end{align}
where $\mathrm{logistic}(x)=(1+\exp(-x))^{-1}$, and
$\mathrm{Bernoulli}(y;p) = p^y (1-p)^{1-y}$ is the Bernoulli distribution with probability of success $p$. Then the posterior log-liklihood can be written as,
\begin{align*}
\log {P(\vw_k|\mathcal{D}_{k-1})} &= -y_k\log(1 + \exp(-\vw_k^\top\phi(\vx_k))) \\
&\qquad - (1-y_k)\log(1+\exp(\vw_k^\top\phi(\vx_k))) \\
&\qquad - \frac{1}{2\sigma^2_d}(\lambda\vw^{\MAP}_{k-1} - \vw_k)^\top(\lambda\vw^{\MAP}_{k-1} - \vw_k)
\end{align*}
Once again, we need to maximize this likelihood to get $\vw^{\MAP}_{k}$.
Taking derivative of this function, we get,
\begin{align*}
\frac{\partial\log {P(\vw_k|\mathcal{D}_{k-1})} }{\partial \vw_k} &= \frac{y_k\exp(-\vw_k^\top\phi(\vx_k))\phi(\vx_k)}{1 + \exp(-\vw_k^\top\phi(\vx_k))} \\
&\qquad - \frac{(1-y_k)\exp(\vw_k^\top\phi(\vx_k))\phi(\vx_k)}{1 + \exp(\vw_k^\top\phi(\vx_k))}  \\
&\qquad - \frac{1}{2\sigma^2_d}(\lambda\vw^{\MAP}_{k-1} - \vw_k).
\end{align*}
As in the Poisson case, it can be observed that the stationary point for which the gradient is zero, can be expressed as $\vw_k = \lambda\vw_{k-1}^\MAP + \alpha_k\phi(\vx_k)$. So, we need to only solve for $\alpha_k$ by maximizing
\begin{align*}
J(\alpha_k) &= -y_k\log(1 + \exp(-\psi_k)\exp(-\alpha_k)) \\
& \quad - (1-y_k)\log(1+\exp(\psi_k)\exp(\alpha_k)) - \frac{\alpha_k^2}{2\sigma^2_d}
\end{align*}
where $\psi_k = \lambda(\vw^{\MAP}_{k-1})^\top\phi(\vx_k)$. The cost function $J(\alpha_k)$ is again strictly concave, and thus, the optimization problem can be solved efficiently. In Fig.~\ref{fig:tracking:logistic}, we demonstrate the effectiveness of this approach in tracking a shifting binary classification boundary.

It should be noted that both Poisson-KLMS and Bernoulli-KLMS do not have an explicit learning rate parameter. This is because the observation model is not Gaussian, where the learning rate parameter is simply the ratio between the diffusion variance and the noise variance. However, in these extensions the diffusion variance plays a similar role; the lower the $\sigma_d^2$, the slower the adaption process.

\section{Experiment}

We acquired data from a wireless communication test bed that is used to evaluate the performance of digital communication systems in realistic indoor environments. The platform is composed of several transmit and receive nodes, each one including a radio-frequency front-end and baseband hardware for signal generation and acquisition. The front-end also incorporates a programmable variable attenuator to control the transmit power value and therefore the signal saturation. A more detailed description of the test bed can be found in \cite{gutierrez2011frequency}. Using the hardware platform, we transmitted clipped orthogonal frequency-division multiplexing (OFDM) signals centered at $5.4$ GHz over real frequency-selective and time-varying channels, with normalized Doppler frequency around $10^{-3}$. The transmit amplifier was operated close to saturation. In this experiment the transmitted and received signals are used to track the variations of the nonlinear channel.

\begin{figure}[t!]
\centering
\includegraphics[width=\columnwidth]{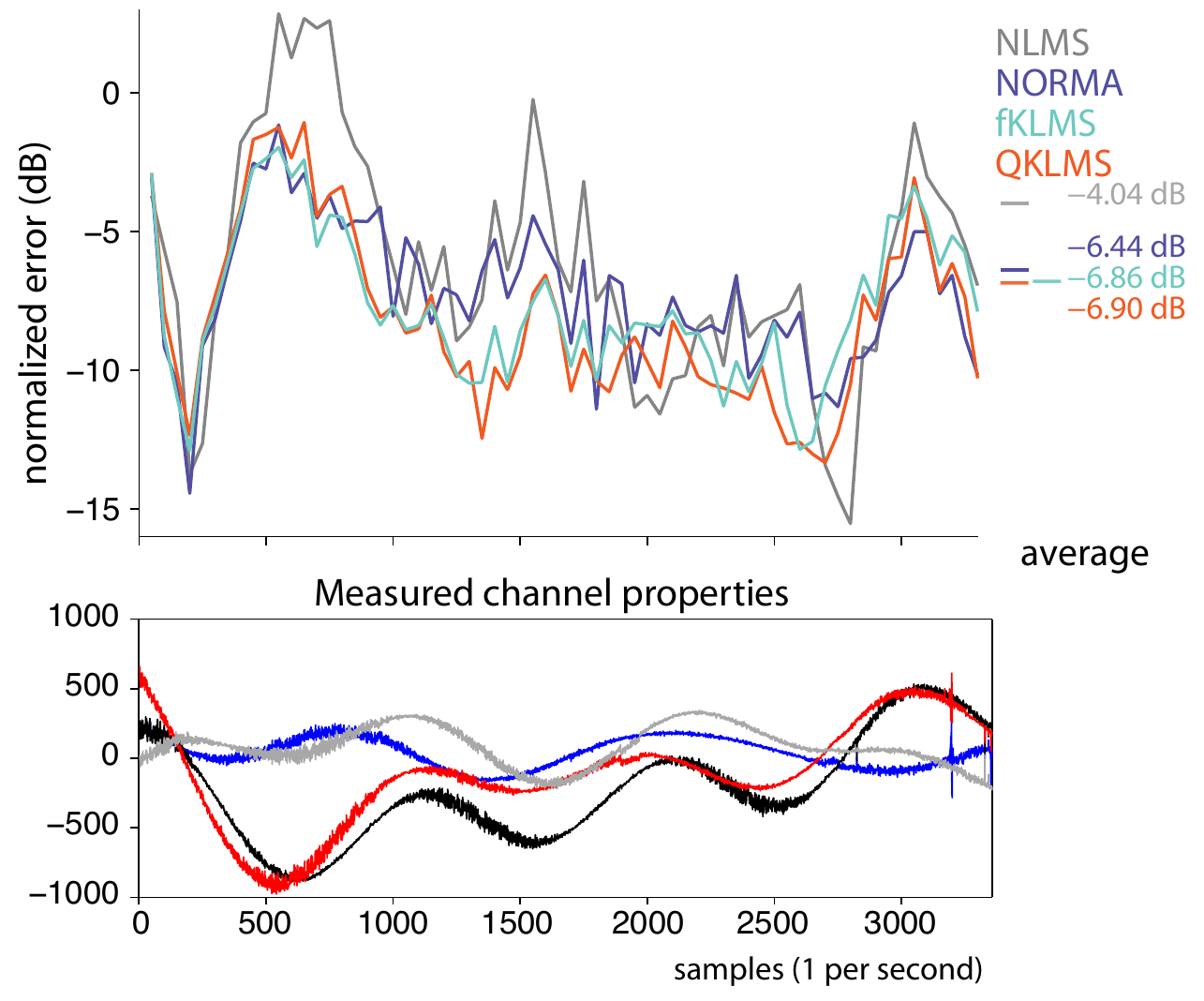}
\caption{(Top) Tracking results on a nonlinear Rayleigh fading channel, using data measured on a test bed with fast time-varying channels.
(Bottom)
Time varying properties of a 4-channel MIMO system.
Real part of the linear stage is measured.
Note that the nonlinearity is weak when the values are close to 0.
}
\label{fig:tracking:testbed}
\end{figure}

We compare 4 algorithms with hyperparameters tuned using the first 500 samples.
Fig.~\ref{fig:tracking:testbed} displays the one-step ahead prediction normalized mean squared error (NMSE) of the tracking experiment.
Average NMSE are: NLMS $-4.0412$, NORMA $-6.4423$, QKLMS $-6.9060$, fKLMS $-6.8624$~dB.
fKLMS and NORMA used 500 total basis functions.
QKLMS and fKLMS show similar performances better than NORMA and (linear) NLMS.

\section{Discussion} \label{sec:conclusion}
In this paper, we derived a family of linear time and space complexity KAF algorithms from Bayesian filtering by maintaining only the MAP solution at each iteration and discarding the posterior distribution (summarized by the covariance).
One of the basic resulting algorithms is identical to KLMS, which is simple and practical.
The tracking ability of LMS/KLMS is usually understood by its stochastic nature that allows it to continually adjust itself to the non-stationary environment.
We provide an alternate explanation of this mechanism by showing the KLMS can also be seen as an approximation to state-space modeling which possesses explicit tracking abilities.
Our framework allows flexibility in the state-space models which can be used
to induce forgetting behavior, as well as novel observation noise models, such as Poisson and Bernoulli.

The optimal nonlinear Bayesian filtering for Gaussian diffusion dynamics, given by either Eq. ~\eqref{eq:diffusion:pure} or Eq. \eqref{eq:OU}, and a Gaussian observation model \eqref{eq:likelihood:gaussian} can be iteratively solved by extended kernel recursive least squares algorithm~\cite{liu2009extended,vanvaerenbergh2012kernel}.
For a general dynamics and a general likelihood such as \eqref{eq:likelihood:poisson}, or \eqref{eq:likelihood:bernoulli}, we often do not have a closed form iterative solution, and one must resort to slower sampling based inference such as sequential Monte Carlo algorithm or approximate inferences.
Exact optimal solutions are impractical due to its high computational cost.
Hence, certain approximations must be made; for example, \cite{vanvaerenbergh2012kernel} uses a low-dimensional subspace approximation.

We have derived KAFs for natural number and binary observation, and it can be also extended to multi-class by using multinomial-logistic model.
However, we do not have convergence results for Poisson and logistic variants as in the Gaussian version where it can be shown that for $\eta<1$ the solution converges in mean.

Our algorithms have a few hyperparameters that need to be set ahead; the kernel parameters, the diffusion variance, and the likelihood parameters.
We have not presented a formal way for choosing those hyperparameters, which plays a crucial role in the speed of convergence and the tracking ability of the algorithm.
If training time series is available, one can use expectation-maximization or sampling methods to find appropriate parameters~\cite{Rasmussen2005}.
We leave these as future work.


\end{document}